%% file: acm/main.tex
\documentclass[sigconf]{acmart}
\AtBeginDocument{%
  }

\setcopyright{acmlicensed}
\copyrightyear{2018}
\acmYear{2018}
\acmDOI{XXXXXXX.XXXXXXX}
\acmConference[Conference acronym 'XX]{conference}{June 03--05,
  2018}{Woodstock, NY}
\acmISBN{978-1-4503-XXXX-X/2018/06}

\usepackage{enumitem}
\usepackage{threeparttable}
\usepackage{pifont}
\usepackage{algorithm}
\usepackage{algorithmic}
\usepackage{mathtools}
\usepackage{booktabs}
\usepackage{multirow}
\usepackage{colortbl}
\usepackage[table]{xcolor}
\usepackage{makecell}
\usepackage{tabularx}

\begin{document}

\title{
\textit{MemAdapter}: Fast Alignment across Agent Memory Paradigms via Generative Subgraph Retrieval}


\author{Xin Zhang}
\affiliation{%
  \institution{The University of Manchester}
  \country{United Kingdom}
}

\author{Kailai Yang}
\affiliation{%
  \institution{The University of Manchester}
  \country{United Kingdom}
}
\email{klyang990203@gmail.com}
\authornote{Corresponding author: Kailai Yang.}

\author{Chenyue Li}
\affiliation{%
  \institution{Stanford University}
  \country{United States}
}

\author{Hao Li}
\affiliation{%
  \institution{Imperial College London}
  \country{United Kingdom}
}

\author{Qiyu Wei}
\affiliation{%
  \institution{The University of Manchester}
  \country{United Kingdom}
}

\author{Jun'ichi Tsujii}
\affiliation{%
  \institution{National Institute of Advanced Industrial Science and Technology}
  \country{Japan}
}

\author{Sophia Ananiadou}
\affiliation{%
  \institution{The University of Manchester}
  \country{United Kingdom}
}
\email{sophia.ananiadou@manchester.ac.uk}

\renewcommand{\shortauthors}{xx et al.}

\begin{abstract}
Memory mechanism is a core component of LLM-based agents, enabling reasoning and knowledge discovery over long-horizon contexts. Existing agent memory systems are typically designed within isolated paradigms (e.g., explicit, parametric, or latent memory) with tightly coupled retrieval methods that hinder cross-paradigm generalization and fusion. In this work, we take a first step toward unifying heterogeneous memory paradigms within a single memory system. We propose \textbf{MemAdapter}, a memory retrieval framework that enables fast alignment across agent memory paradigms. MemAdapter adopts a two-stage training strategy: (1) training a \textit{generative subgraph retriever} from the \textit{unified memory space}, and (2) adapting the retriever to unseen memory paradigms by training a lightweight \textit{alignment module} through contrastive learning. This design improves the flexibility for memory retrieval and substantially reduces alignment cost across paradigms. Comprehensive experiments on three public evaluation benchmarks demonstrate that the generative subgraph retriever consistently outperforms five strong agent memory systems across three memory paradigms and agent model scales. Notably, MemAdapter completes cross-paradigm alignment within 13 minutes on a single GPU, achieving superior performance over original memory retrievers with less than 5\% of training compute. Furthermore, MemAdapter enables effective zero-shot fusion across memory paradigms, highlighting its potential as a plug-and-play solution for agent memory systems. This project will be publicly available on \href{https://github.com/xstarx1212/MemAdapter}{\textcolor{blue}{Github}}.
\end{abstract}



\maketitle

\input{acm/Sections/intro}

\input{acm/Sections/method}

\input{acm/Sections/experiments}
\input{acm/Sections/related_work}
\input{acm/Sections/conclusion}


\bibliographystyle{ACM-Reference-Format}
\bibliography{custom}

\appendix
\input{acm/Sections/appn}

\end{document}

%% file: acm/Sections/intro.tex
\section{Introduction}
\begin{figure*}[t!]
  \centering
  \includegraphics[scale=0.47]{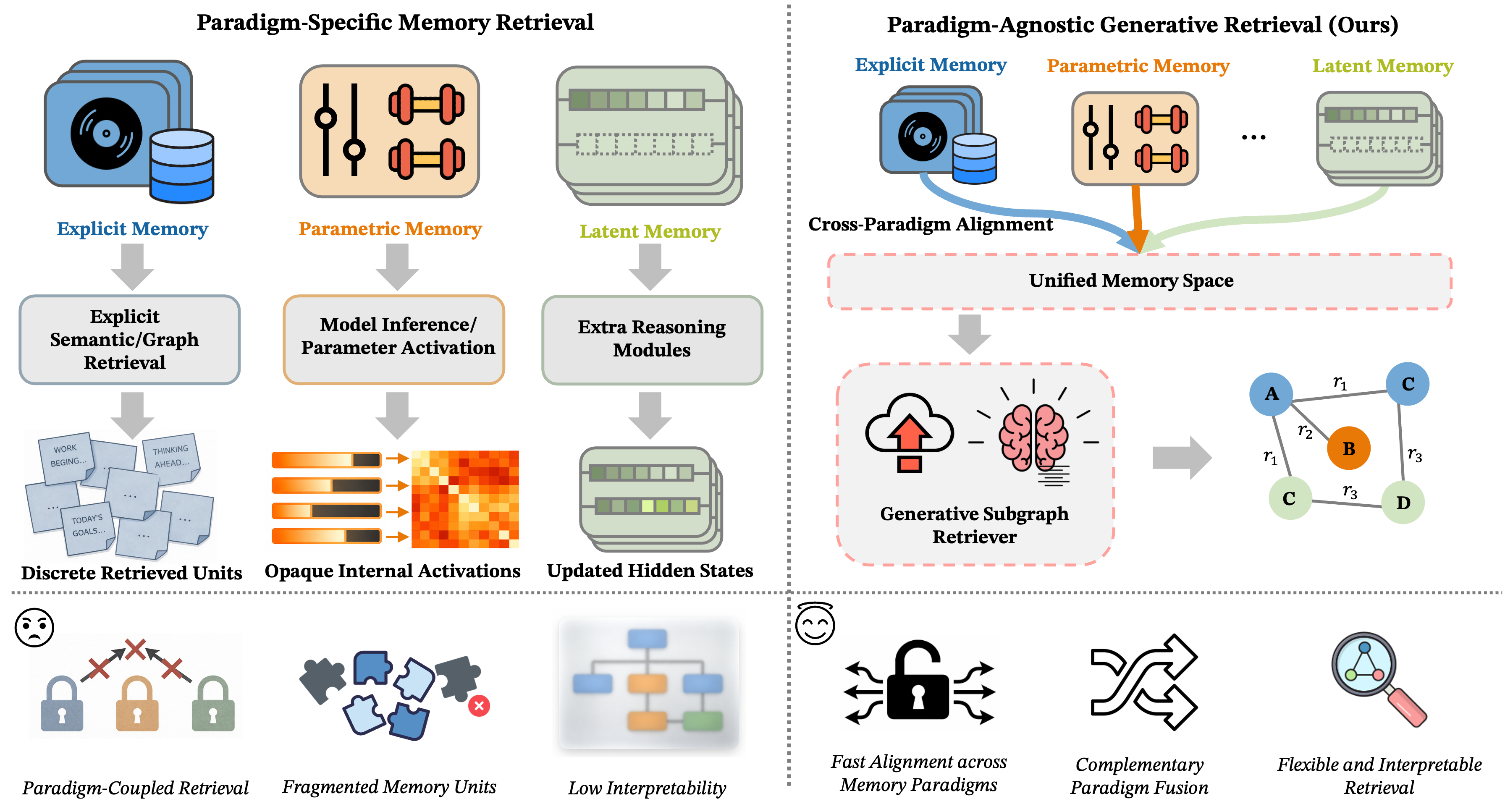}
\caption{
Comparison between paradigm-specific memory retrieval and paradigm-agnostic generative retrieval.
}
  \label{fig:intro_paradigm}
\end{figure*}

Memory mechanisms have emerged as a foundational component in the design of modern large language model (LLM)-based agent systems~\cite{zhang2025survey,hu2025memory,jiaai}. They enable agents to accumulate, retain, and discover complex information and knowledge beyond the limitations imposed by fixed context windows and single-step interactions with the environment. Recent taxonomies~\cite{hu2025memory} broadly categorize agent memory into three primary paradigms with consequent highly-coupled memory retrieval designations for downstream applications: (1) \textbf{Explicit memory} represents information as discrete, externally stored units organized in structured forms, such as temporal sequence buffers and hierarchical tree/graph topologies~\cite{NEURIPS2023_1b44b878,kim2025pre,wu2025sgmemsentencegraphmemory}, accompanied by explicit semantic/graph retrieval methods. (2) \textbf{Parametric memory} is embedded within distributed representations of the model parameters and retrieved directly through model inference. This paradigm encompasses memory encoded in the base model parameters through large-scale training~\cite{pouransari2025pretraining,su2025scaling} or externalized parametric memory implemented via auxiliary modules, such as LoRA and adapters~\cite{lin2025continual,yao2023retroformer}. (3) \textbf{Latent memory} is maintained in separate distributed representations and dynamically updated during task execution, such as key–value caches or latent embedding spaces~\cite{zhang2025memgen,zhang2023h2o}. Relevant information is usually infused through extra trainable modules, such as attention and reasoning mechanisms. The above paradigms emphasize different features, such as interpretability and computational efficiency~\cite{zhang2026implicit}.

Despite achieving strong performance, the heavy reliance of existing memory retrieval methods on specific agent memory paradigms introduces several key limitations, as illustrated in Figure~\ref{fig:intro_paradigm} (left). First, the tight coupling between memory paradigms and their corresponding retrieval mechanisms substantially constrains cross-paradigm memory alignment and fusion. For example, adapting graph-based memory systems~\cite{wu2025sgmemsentencegraphmemory,edge2024local,rasmussen2025zep} to latent memory settings typically requires extensive architectural redesign and resource-intensive continual training. Such adaptations are further hampered by catastrophic forgetting~\cite{ramasesh2021effect}, which degrades the original graph-based retrieval capabilities and impedes effective fusion between graph-structured and latent memory. Moreover, paradigm-specific retrieval strategies inherently introduce paradigm-dependent weaknesses into the memory system. In explicit memory frameworks, retrieval commonly relies on semantic or topological similarity to select fragmented memory units, thereby limiting flexibility in the compositionality and granularity of retrieved evidence. In parametric and latent memory systems, relevant information is accessed through model- and task-specific parameter activations, resulting in opaque internal representations that are difficult to disentangle, reuse, or interpret.

In light of these limitations, we take a first step toward unifying the major agent memory paradigms via paradigm-agnostic generative retrieval. As shown in Fig~\ref{fig:intro_paradigm} (right), we introduce \textbf{MemAdapter}, a memory retrieval framework designed to enable efficient and fast alignment across memory paradigms (e.g., explicit, parametric, and latent memory). MemAdapter adopts a two-stage training strategy. Firstly, we train a \textit{generative subgraph retriever} via model distillation, which conditions on task information to generate an explicit memory subgraph from a \textit{unified memory space}. By integrating generative retrieval with graph-structured evidence, this design substantially improves both the flexibility and interpretability of memory retrieval. Secondly, lightweight \textit{alignment modules} are trained to achieve adaptation to unseen memory paradigms through contrastive learning~\cite{chen2020simple,gao2021simcse} in the \textit{unified memory space}. This alignment is highly efficient, as it optimizes only a lightweight \textit{alignment module} and achieves effective adaptation via sampling from merely 2,500 demonstrations.

\begin{figure*}[t!]
  \centering
  \includegraphics[scale=0.55]{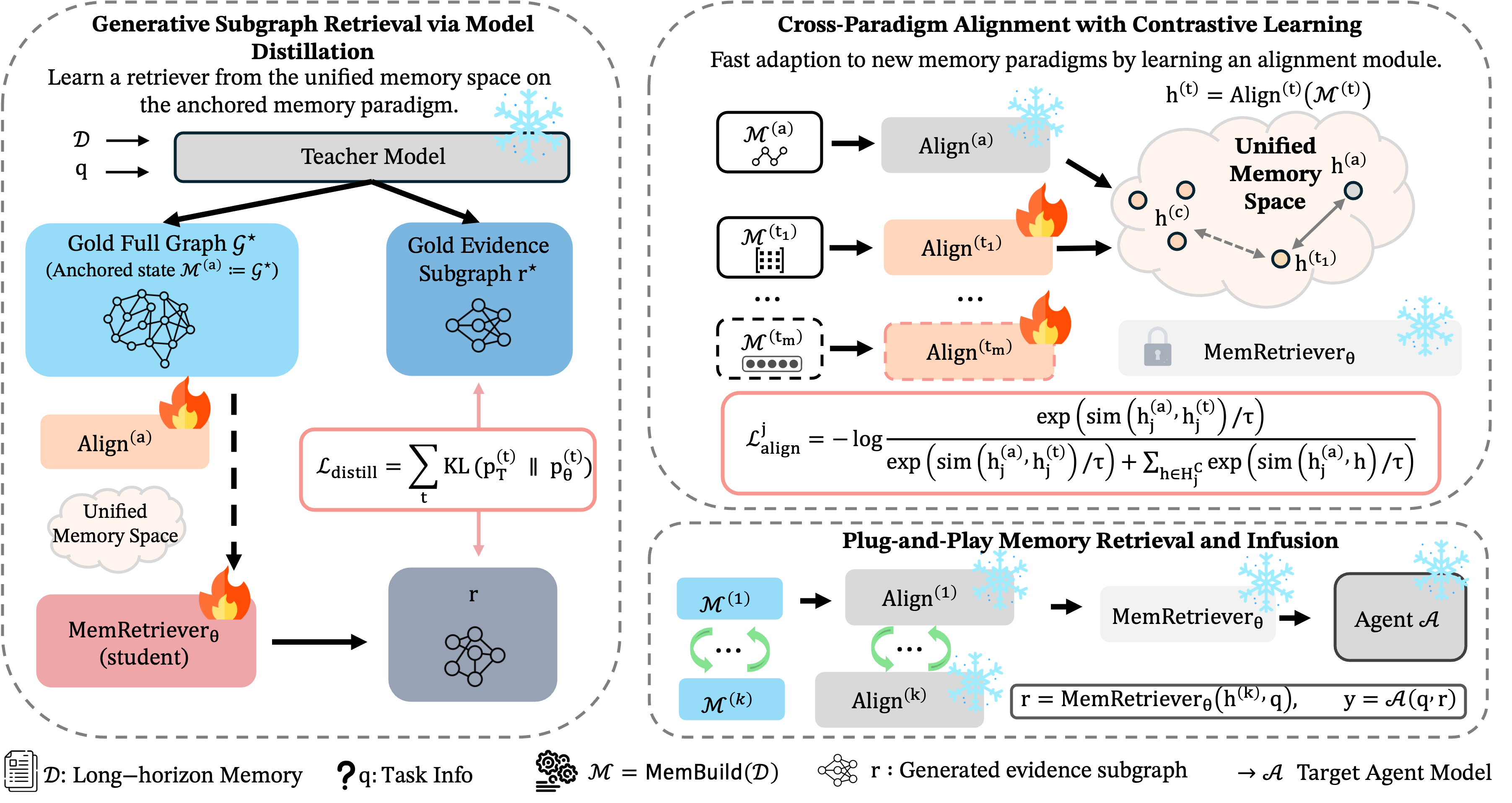}
\caption{
MemAdapter performs generative subgraph retrieval from a \textbf{unified memory space} adapted by paradigm-specific alignment modules. Building on an anchored memory paradigm, the first stage introduces a teacher model to provide imitation-based supervision for the retriever. In stage two, the anchored alignment module is used to supervise a lightweight alignment module for the target paradigm via contrastive learning. At inference, aligned memory paradigms can be used and infused in a plug-and-play manner.
}
  \label{fig:method}
\end{figure*}

Comprehensive evaluation on three memory-intensive question answering benchmarks shows that the generative subgraph retriever consistently outperforms strong baseline methods across explicit, parametric, and latent memory paradigms for agent models with parameter scales ranging from 1.5B to 7B. Notably, MemAdapter attains performance superior to state-of-the-art explicit and implicit memory retrieval approaches after the corresponding alignment stages, which completes within 13 minutes on a single GPU and requires less than 5\% of training-time computational resources compared to tuning the original memory retrievers on the agent models. Furthermore, MemAdapter exhibits strong paradigm-fusion capabilities, as it substantially enhances agent model performance by jointly leveraging complementary information from explicit, parametric, and latent memory. These results highlight MemAdapter as a promising first step toward plug-and-play memory fusion across heterogeneous memory paradigms. In summary, this work makes the following contributions:

\begin{itemize}[leftmargin=8pt]
\item We propose \textbf{MemAdapter}, the first memory retrieval framework that unifies multiple memory paradigms. It trains on a moderate scale to perform generative subgraph retrieval, then achieves highly efficient alignment across paradigms via contrastive learning.

\item MemAdapter outperforms competitive memory systems across three public evaluation benchmarks, three memory paradigms, and agent models scaling from 1.5B to 7B. Notably, it can achieve such performance on a fast cross-paradigm alignment basis with less than 5\% of training compute.

\item MemAdapter performs reliable complementary fusion of explicit, parametric, and latent memory in a zero-shot manner, highlighting it as a promising first step toward plug-and-play memory fusion across heterogeneous paradigms.
\end{itemize}

%% file: acm/Sections/method.tex
\section{Methodology}
\subsection{Agent Memory Systems}
\label{subsec:agent_memory_system}
Agent memory systems aim to support complex reasoning tasks over long-horizon memory beyond current contexts. The system mainly consists of three components: (1) \emph{Memory state} builds memory representations constructed from the long-horizon memory. (2) \emph{Memory retriever} extracts task-specific evidence from the memory state conditioned on the query. (3) \emph{Agent model} produces the final output given the user query and task-relevant evidence provided by the \emph{memory retriever}.

\subsubsection{Memory State}
Memory state $\mathcal{M}$ is a structured representation constructed from the long-horizon memory $\mathcal{D}$,
\begin{equation}
\mathcal{M} = \textit{MemBuild}(\mathcal{D}).
\label{eq:mem_build}
\end{equation}
Memory paradigms usually determine how the building methodology $\textit{MemBuild}$ is instantiated. For example, explicit memory stores externalized units such as text records or graphs, parametric memory encodes information in model parameters, and latent memory maintains information in evolving latent representations.

\subsubsection{Memory Retriever}
Given the target query $q$ and the memory state $\mathcal{M}$, the memory retriever (parameterized by $\theta$) produces the task-specific evidence $r$,
\begin{equation}
r = \textit{MemRetriever}_{\theta}(\mathcal{M}, q),
\label{eq:mem_retrieve}
\end{equation}
The evidence $r$ contains relevant information used by the agent model. In MemAdapter, $r$ is instantiated as an explicit subgraph.

\subsubsection{Agent Model}
Due to the heavy redundancy of long-horizon memory, the agent model $\textit{Agent}$ usually does not directly access $\mathcal{M}$. Instead, it conditions on the task information $q$ and $r$ as the task-relevant memory information to produce the output $y$:
\begin{equation}
y = \textit{Agent}(q, r).
\label{eq:agent}
\end{equation}



\subsection{MemAdapter}

As shown in Figure~\ref{fig:method}, MemAdapter follows a two-stage training procedure. Firstly, the generative subgraph retriever is trained via model distillation. We then perform cross-paradigm alignment via contrastive learning. During inference, memory states are efficiently infused in a plug-and-play manner.

\subsubsection{Generative Subgraph Retrieval via Model Distillation}
\label{sec:stage1_distill}
In this stage, we aim to train a generative subgraph retriever to construct structured evidence subgraphs based on the task information and memory from a distributed \textit{unified memory space} $\mathcal{S}\in\mathbb{R}^{D_s}$, where $D_s$ denotes the space dimension. The \textit{unified memory space} is designed to be independent of memory paradigms, a key step towards plug-and-play memory retrieval. During training, a central challenge is the absence of large-scale ground-truth evidence subgraphs for supervision. Moreover, self-supervised learning via in-context learning of the retriever itself can bring inductive biases and be unstable across instances and tasks. We adopt a model distillation framework that constructs imitable supervision directly from a strong teacher model.

We start by defining an anchored memory paradigm $a$, which is instantiated as \textit{explicit graph-based memory}, but can be replaced by any existing memory paradigms. Specifically, the teacher model is first prompted to construct a full memory graph $\mathcal{G}^{\star}$ from the memory $\mathcal{D}$. 
We utilize the generated graph as the input and instantiate Eqn.~\ref{eq:mem_build} with state-of-the-art method~\cite{zhang2026implicit} for memory state construction:
\begin{equation}
\mathcal{M}^{(a)} \coloneqq \textit{MemBuild}(\mathcal{G}^{\star})
\label{eq:anchor_memory_state}
\end{equation}

To achieve alignment between the anchored memory paradigm and the \textit{unified memory space}, we introduce an anchored alignment module $\mathsf{Align}^{(a)}$ that maps the anchored memory state $\mathcal{M}^{(a)}$ into the unified memory space:
\begin{equation}
h^{(a)} = \mathsf{Align}^{(a)}\!\left(\mathcal{M}^{(a)}\right).
\label{eq:anchor_align}
\end{equation}
where $h^{(a)}\in\mathcal{S}$ denotes the memory representation of $\mathcal{M}^{(a)}$ within the \textit{unified memory space}, and $\mathsf{Align}^{(a)}$ is parameterized with a lightweight neural network.

We formulate memory retrieval as a conditional auto-regressive generation process of a structured evidence subgraph, based on the task information $q$ and the anchored memory representation $h^{(a)}$. The generated subgraph $r$ is expected to aggregate relevant memory units and determine their appropriate relations flexibly. Formally, the memory retriever $\mathsf{MemRetrieve}_{\theta}$ produces the evidence subgraph as follows:
\begin{equation}
r = \mathsf{MemRetrieve}_{\theta}\!\left(h^{(a)}, q\right)
\label{eq:retriever_dist}
\end{equation}

During training, the teacher model is further prompted to produce a query-conditioned evidence subgraph $r^{\star}$, constrained to be a subgraph of $\mathcal{G}^{\star}$. $r^{\star}$ is serialized into token sequences with a structure-preserving linearization scheme to enable autoregressive supervision. More details about the scheme are shown in Appendix \ref{sec:appendix_teacher}.
We optimize the student model by directly minimizing a token-level distribution distillation objective:
\begin{equation}
\mathcal{L}_{\mathrm{distill}}=\frac{1}{|r^{\star}|}\sum_{t=1}^{|r^{\star}|}\mathrm{KL}\left(p_{T}(r_t|r_{<t},q, h^{(a)})\;\|\;p_{\theta}(r^{\star}_t|r^{\star}_{<t},q,\mathcal{G}^{\star})\right).
\label{eq:distill_loss}
\end{equation}
where $p_{T}(\cdot)$ and $p_{\theta}(\cdot)$ denote the teacher and student models' token distributions, and $KL(\cdot||\cdot)$ computes the KL-divergence between two distributions. This objective enables the generative subgraph retriever to imitate the teacher model's behavior in a fine-grained manner.

\input{acm/Tables/CL_algo}

\subsubsection{Cross-Paradigm Alignment with Contrastive Learning}
\label{sec:stage2_align}
This stage aims to efficiently adapt the generative subgraph retriever to unseen memory paradigms. Such adaptation is nontrivial since heterogeneous memory states differ in structure and storage, yielding significantly degraded retrieval performance due to the mismatch of patterns in the \textit{unified memory space} with the anchored memory paradigm. We resolve this mismatch by achieving fast alignment between the anchored memory paradigm and the current target paradigm $t$. Specifically, we train a lightweight paradigm-specific alignment module via contrastive learning on merely 2,500 demonstrations in the \textit{unified memory space}. Firstly, the anchored alignment module and the target alignment module $\mathsf{Align}^{(t)}_{\lambda}$ map the corresponding memory states into the \textit{unified memory space}:
\begin{equation}
h^{(a)} = \mathsf{Align}^{(a)}\!\left(\mathcal{M}^{(a)}\right),
\qquad
h^{(t)} = \mathsf{Align}^{(t)}_{\lambda}\!\left(\mathcal{M}^{(t)}\right),
\label{eq:aligned_repr_stage2}
\end{equation}
where $h^{(a)},h^{(t)}\in\mathcal{S}$, $\mathsf{Align}^{(a)}$ is fixed, and $\mathsf{Align}^{(t)}_{\lambda}$ is a trainable module parameterized with another lightweight neural network.

As depicted in Algorithm~\ref{alg:alignment}, $\mathsf{Align}^{(t)}_{\lambda}$ is optimized with a simple yet effective contrastive learning objective that enforces instance-level alignment between the anchored and target memory paradigms. For the $j$-th instance sampled from the demonstration set, we randomly select a negative set $H^C_j$, compute the corresponding representations, and minimize an InfoNCE-like loss:
\begin{equation}
\mathcal{L}^{j}_{\mathrm{align}}
=
- \log
\frac{
\exp\!\left(\mathrm{sim}(h^{(a)}_j, h^{(t)}_j)/\tau\right)
}{
\exp\!\left(\mathrm{sim}(h^{(a)}_j, h^{(t)}_j)/\tau\right)
+
\sum\limits_{h \in H^C_j}
\exp\!\left(\mathrm{sim}(h^{(a)}_j, h)/\tau\right)
},
\label{eq:contrastive_loss}
\end{equation}
where $\mathrm{sim}(\cdot,\cdot)$ denotes cosine similarity and $\tau$ is the temperature parameter. For each target memory paradigm, we separately optimize its alignment module $\mathsf{Align}^{(t)}_{\lambda}$ using the contrastive objective in Eq.~\eqref{eq:contrastive_loss}. These alignment modules ensure that heterogeneous memory states are appropriately projected into the \textit{unified memory space} to ensure correct generative subgraph retrieval.

\input{acm/Tables/main_results}

\subsection{Plug-and-Play Memory Retrieval and Fusion}
\label{sec:infusion}
MemAdapter enables a unified generative subgraph retrieval process for heterogeneous memory paradigms without customized architectures for the memory retriever and agent model. Any new memory paradigms can be efficiently adapted to MemAdapter by training a lightweight alignment module via the procedure in Sec.~\ref{sec:stage2_align}, which normally finishes on a single GPU within 13 minutes. The alignment module can then be used on any memory states produced under the same paradigm. Changing memory paradigm only requires switching to the new alignment module to obtain memory representations, which enables plug-and-play memory retrieval for any memory paradigms and agent systems. This flexibility empowers agent memory systems in many practical scenarios, including supporting agent reasoning with heterogeneous memory states~\cite{hu2025memory,DBLP:journals/corr/abs-2507-07957} and cross-paradigm knowledge infusion in a cold-start manner~\cite{DBLP:journals/corr/abs-2502-06872}.

MemAdapter also marks the first step towards plug-and-play heterogeneous memory fusion, which leverages memory from multiple paradigms to support the same target agent model system. Specifically, $k$ memory states from different paradigms: 
\{$\mathcal{M}_0^{(t_1)}$, $\mathcal{M}_1^{(t_1)}$, $\mathcal{M}_2^{(t_2)}$, ... $\mathcal{M}_k^{(t_m)}$\}
are first projected into the \textit{unified memory space} via their corresponding alignment modules:
[$h_0^{(t_1)}$, $h_1^{(t_1)}$, $h_2^{(t_2)}$, ... ,$h_k^{(t_m)}$]
These aligned representations can be easily fused via any down-sampling techniques:
\begin{equation}
\hat{h} = \mathsf{DownSampling}\!\left(h_0^{(t_1)},h_1^{(t_1)},...,h_k^{(t_m)}\right)
\label{eq:downsample}
\end{equation}
where $\hat{h}\in\mathcal{S}$ and we instantiate a simple max-pooling method for down-sampling. $\hat{h}$ is used in Eqn. \ref{eq:retriever_dist} as memory information to obtain the retrieved memory subgraph $\hat{r}$, which explicitly combines task-relevant information from heterogeneous memory states and is used to support the agent model as in Eqn. \ref{eq:agent}.

%% file: acm/Tables/CL_algo.tex
\begin{algorithm}[h]
\caption{Cross-paradigm alignment via contrastive learning in the unified memory space.}
\label{alg:alignment}
\begin{algorithmic}[1]
\REQUIRE The demonstration number $N$, Negative sample size $C$, batch size $B$, the anchored memory states $\{\mathcal{M}_i^{(a)}\}_{i=1}^N$, target-paradigm memory states $\{\mathcal{M}_i^{(t)}\}_{i=1}^N$, the anchored alignment module $\textit{Align}^{(a)}$.
\ENSURE Alignment module $\textit{Align}^{(t)}_\lambda$ for the target paradigm $t$.
\STATE Randomly initialize $\textit{Align}^{(t)}_\lambda$ with parameters $\lambda$.
\STATE Freeze the model parameters for $\mathrm{Align}^{(a)}$.
\FOR{each training iteration}
    \STATE Randomly sample a minibatch with size $B$: $\{\mathcal{M}_j^{(a)}\}_{j=1}^B$ and $\{\mathcal{M}_j^{(t)}\}_{j=1}^B$ from $\{\mathcal{M}_i^{(a)}\}_{i=1}^N$ and $\{\mathcal{M}_i^{(t)}\}_{i=1}^N$
    \FOR{each $\mathcal{M}_j^{(a)}\in\{\mathcal{M}_j^{(a)}\}_{j=1}^B$ and $\mathcal{M}_j^{(t)}\in\{\mathcal{M}_j^{(t)}\}_{j=1}^B$}
        \STATE Compute $h_j^{(a)}=\textit{Align}^{(a)}(\mathcal{M}_j^{(a)})$
        \STATE Compute $h_j^{(t)}=\textit{Align}^{(t)}_\lambda(\mathcal{M}_j^{(t)})$
        \STATE Randomly sample a negative sets with size $C$: $\{\mathcal{M}_k^{(a)}\}_{k=1}^C$ from $\{\mathcal{M}_i^{(a)}\}_{i=1}^N\setminus \{\mathcal{M}_j^{(a)}\}$
        \STATE Initialize a set $H^C_j=\varnothing$
        \FOR{each $\mathcal{M}_k^{(a)}\in\{\mathcal{M}_k^{(a)}\}_{k=1}^C$}
            \STATE Compute $h_k^{(c)}=\textit{Align}^{(a)}(\mathcal{M}_k^{(a)})$
            \STATE $H^C_j\cup \{h_k^{(c)}\}$
        \ENDFOR
        \STATE Compute $\mathcal{L}_{\mathrm{align}}^j=\mathcal{L}_{\mathrm{align}}(h_j^{(a)}, h_j^{(t)},H^C_j)$ using Eq.~\eqref{eq:contrastive_loss}
    \ENDFOR
    \STATE Optimize $\textit{Align}^{(t)}_\lambda$ via gradient: $\nabla_{\lambda}\frac{1}{B}\sum_{j=1}^{B}\mathcal{L}_{\mathrm{align}}^{j}$
\ENDFOR
\end{algorithmic}
\end{algorithm}

%% file: acm/Tables/main_results.tex
\begin{table*}[t]
\centering
\small
\setlength{\tabcolsep}{4.4pt}
\renewcommand{\arraystretch}{1.18}

\definecolor{dshead}{RGB}{238,238,238}
\definecolor{bboneA}{RGB}{245,248,255}
\definecolor{bboneB}{RGB}{246,252,246}
\definecolor{bboneC}{RGB}{255,248,240}

\caption{
Main results on three QA benchmarks. In this case, we report MemAdapter performance by the generative subgraph retriever trained on the anchored memory paradigm. The results are grouped by the scale of agent models, and all methods are evaluated under their best reported configurations with memory enabled.
}
\label{tab:main_results}

\begin{tabularx}{\textwidth}{l *{12}{>{\centering\arraybackslash}X}}
\toprule
\multirow{2}{*}{\textbf{Method}} &
\multicolumn{3}{c}{\cellcolor{dshead}\textbf{WikiMultiHopQA}} &
\multicolumn{3}{c}{\cellcolor{dshead}\textbf{NarrativeQA}} &
\multicolumn{3}{c}{\cellcolor{dshead}\textbf{MuSiQue}} &
\multicolumn{3}{c}{\cellcolor{dshead}\textbf{Avg}} \\
\cmidrule(lr){2-4}\cmidrule(lr){5-7}\cmidrule(lr){8-10}\cmidrule(lr){11-13}
& \textbf{EM} & \textbf{F1} & \textbf{R-1}
& \textbf{EM} & \textbf{F1} & \textbf{R-1}
& \textbf{EM} & \textbf{F1} & \textbf{R-1}
& \textbf{EM} & \textbf{F1} & \textbf{R-1} \\
\midrule

\multicolumn{13}{c}{\cellcolor{bboneA}\textbf{Agent Model: Qwen2.5-1.5B-Instruct}} \\
A-Mem &
8.31 & 17.02 & 18.94 &
7.21 & 24.37 & 27.46 &
4.68 & 9.88 & 10.41 &
6.92 & 16.27 & 18.03 \\

Mem0  &
\underline{12.04} & 19.21 & 20.87 &
\underline{9.73} & \underline{30.42} & \underline{32.18} &
\underline{7.21} & \underline{14.73} & \underline{16.88} &
\underline{11.63} & \underline{21.94} & \underline{23.11} \\

MemoryLLM &
7.58 & 12.96 & 14.42 &
4.63 & 26.71 & 30.02 &
1.34 & 3.67 & 3.81 &
4.41 & 13.88 & 15.02 \\

StreamingLLM &
9.94 & \underline{20.63} & \underline{21.48} &
6.18 & 28.12 & 31.57 &
5.26 & 10.01 & 12.17 &
8.54 & 18.92 & 21.36 \\

CARE &
0.47 & 6.18 & 6.52 &
0.73 & 8.87 & 9.14 &
0.56 & 4.83 & 5.07 &
0.59 & 6.63 & 6.91 \\

\textbf{MemAdapter (Ours)} &
\textbf{14.08} & \textbf{23.03} & \textbf{23.75} &
\textbf{13.30} & \textbf{36.00} & \textbf{37.83} &
\textbf{14.36} & \textbf{22.84} & \textbf{23.80} &
\textbf{13.91} & \textbf{27.29} & \textbf{28.46} \\
\multicolumn{13}{c}{\cellcolor{bboneB}\textbf{Agent Model: Qwen2.5-3B-Instruct}} \\
A-Mem &
14.00 & 23.99 & 24.49 &
20.65 & 39.62 & 41.42 &
8.44 & 15.89 & 16.55 &
14.36 & 26.50 & 27.49 \\

Mem0 &
\underline{18.73} & \underline{28.56} & \underline{29.27} &
\underline{21.81} & \underline{43.68} & \underline{45.24} &
\underline{14.94} & \underline{24.36} & \underline{25.25} &
\underline{18.49} & \underline{32.20} & \underline{33.25} \\

MemoryLLM &
12.10 & 19.73 & 20.40 &
14.24 & 42.57 & 44.42 &
2.44 & 5.86 & 6.52 &
9.59 & 22.72 & 23.78 \\

StreamingLLM &
15.91 & 26.01 & 26.91 &
15.28 & 42.77 & 44.61 &
8.03 & 15.98 & 16.58 &
13.07 & 28.25 & 29.37 \\

CARE &
0.69 & 7.41 & 7.86 &
0.94 & 10.32 & 10.78 &
0.63 & 5.26 & 5.49 &
0.75 & 7.66 & 8.04 \\

\textbf{MemAdapter (Ours)} &
\textbf{19.12} & \textbf{29.99} & \textbf{30.81} &
\textbf{30.25} & \textbf{55.35} & \textbf{56.99} &
\textbf{21.93} & \textbf{34.81} & \textbf{36.38} &
\textbf{23.77} & \textbf{40.05} & \textbf{41.39} \\

\midrule

\multicolumn{13}{c}{\cellcolor{bboneC}\textbf{Agent Model: Qwen2.5-7B-Instruct}} \\
A-Mem &
12.87 & 23.53 & 24.33 &
22.13 & 41.51 & 43.16 &
13.74 & 23.00 & 23.53 &
16.25 & 29.35 & 30.34 \\

Mem0 &
8.83 & 17.92 & 18.05 &
26.67 & 50.01 & 51.62 &
\underline{23.33} & \underline{36.73} & \underline{37.72} &
\underline{22.32} & \underline{38.75} & \underline{39.66} \\

MemoryLLM &
12.69 & 19.91 & 20.43 &
\underline{29.51} & \underline{56.91} & \underline{58.73} &
3.48 & 7.34 & 8.35 &
15.23 & 28.05 & 29.17 \\

StreamingLLM &
\underline{18.49} & \underline{29.27} & \underline{30.10} &
26.67 & 50.01 & 51.62 &
18.74 & 29.56 & 30.32 &
21.30 & 36.28 & 37.35 \\

CARE &
0.98 & 8.61 & 9.03 &
1.24 & 12.71 & 13.18 &
0.86 & 6.17 & 6.42 &
1.03 & 9.16 & 9.58 \\

\textbf{MemAdapter (Ours)} &
\textbf{19.28} & \textbf{29.73} & \textbf{30.38} &
\textbf{34.79} & \textbf{61.59} & \textbf{63.22} &
\textbf{27.64} & \textbf{40.20} & \textbf{41.54} &
\textbf{27.24} & \textbf{43.84} & \textbf{45.05} \\

\bottomrule
\end{tabularx}
\end{table*}

%% file: acm/Sections/experiments.tex
\section{Experiments}

\input{acm/Sections/experimental_setting}

\subsection{Main Results}
\label{sec:main_results}

We compare MemAdapter trained under the anchored memory paradigm with the baseline agent memory systems. The results are summarized in Table~\ref{tab:main_results}. The generative subgraph retriever consistently achieves the best overall performance across benchmarks and agent model scales, significantly outperforming baseline systems in terms of average EM, F1, and ROUGE-1 under the 1.5B, 3B, and 7B settings. These results demonstrate the effectiveness of generative subgraph retrieval for agent memory. In contrast, explicit memory systems such as A-Mem and Mem0 exhibit limited performance across model scales, as they primarily rely on similarity-based retrieval over fixed memory units. Parametric and latent memory systems show more oscillated performance across tasks and model sizes, reflecting their reliance on paradigm-specific retrieval mechanisms and the internal reasoning capacity of the underlying agent models.

These results also highlight benchmark-dependent failure cases that help clarify the limitations of different memory paradigms. For example, StreamingLLM performs competitively on WikiMultiHopQA and NarrativeQA, but its performance drops substantially on MuSiQue across all agent model scales. StreamingLLM is designed to support efficient long-context inference through attention sink and cache mechanisms, rather than to retrieve task-relevant evidence explicitly. In contrast, MuSiQue requires preserving sparsely distributed evidence across multi-step reasoning chains, making it highly sensitive to the removal of intermediate information. As a result, partial memory truncation can lead to severe information loss and degraded performance. MemAdapter mitigates this issue by constructing an explicit, task-relevant evidence subgraph, which preserves critical dependencies necessary for multi-step reasoning and improves robustness across benchmarks.

As the agent model scales, MemAdapter consistently outperforms parametric and latent memory systems, even though the performance of these baselines generally improves with larger models. Notably, when paired with the strong Qwen2.5-7B-Instruct agent model, MemAdapter still surpasses competitive parametric baselines across all three benchmarks. They indicate that MemAdapter provides complementary, task-relevant information that extends beyond the intrinsic capacity of the agent model, using only lightweight alignment modules and a generative subgraph retriever. Moreover, they suggest that leveraging the in-context learning capabilities of agent models for memory-dependent reasoning is an effective and scalable approach.

\input{acm/Tables/add_memadapter}

\subsection{Cross-paradigm Alignment Performance}
We select one representative memory system from each paradigm and conduct alignment training to project their memory states into the unified memory space, facilitating retrieval using the generative subgraph retriever. MemAdapter completes cross-paradigm alignment \textbf{with less than 5\% of training compute compared to tuning the original memory retrievers on the agent models}. We present the performance comparisons in Table~\ref{tab:memadapter_effect}.

The results show that MemAdapter consistently outperforms the original memory retrievers in downstream performance across memory paradigms and agent model scales. The performance gains are particularly significant on the latent memory system MemoryLLM, where MemAdapter shows an average of 7.63\% advantage in R-1 across all agent models. These results show that converting implicit memory states into explicit evidence subgraphs can lead to substantial improvements in agent model performance and improve the interpretability of the memory system. The explicit memory system A-Mem also benefits from using MemAdapter, with stable performance gains across all agent model scales. This is because generative subgraph retrieval can further refine similarity-based retrieval results to provide more flexible and accurate memory information to the agent model. For StreamingLLM, the performance of MemAdapter is less significant compared to other memory systems, possibly because StreamingLLM does not perform any memory refinement before producing memory states and mostly relies on the agent model in memory retrieval. The noisy information makes the projection more challenging for the alignment module. Interestingly, at 7B scale, MemAdapter slightly reduces F1 but markedly increases ROUGE-1, suggesting a shift toward more evidence-grounded and lexically coherent generation. 
Overall, these results show that MemAdapter can effectively enhance the utilization of heterogeneous memory through generative subgraph retrieval from a unified memory space. Adapting MemAdapter to new memory paradigms with a fast alignment process is also much more efficient than optimizing the agent model for the original memory retrieval methods, which saves over 95\% of training compute.

\input{acm/Tables/memory_quality}

\subsection{Memory Efficiency and Utilization}
\label{sec:memory_quality}
We analyze memory efficiency and utilization on 500 randomly selected samples from the benchmarks. 
We use retrieved memory text for explicit memory and the full context for parametric and latent systems. For MemAdapter, we use the generated evidence subgraph. 
As shown in Table~\ref{tab:memory_quality}, \textbf{Mem Length} measures the average length of the memory in characters. The \textbf{Unique Ratio} reflects the proportion of unique tokens. \textbf{Memory Utilization} computes the ratio between the correctly-answered samples and the total gold samples.

The results show that MemAdapter reduces Mem Length by more than 50\% compared with all baseline methods. Parametric and latent memory systems require the agent model to process the full memory, with an average length exceeding 5.8K characters, which substantially increases the reasoning burden. Explicit memory systems achieve a notable reduction in memory length but still retain an average length of over 4.3K characters.
In contrast, MemAdapter produces a compact, task-conditioned evidence subgraph, reducing the average memory length to 2.2K characters. This result demonstrates its strong capability for effective information compression. MemAdapter also attains a comparable unique ratio to other explicit memory systems. Considering its significantly lower mem length, this indicates its higher efficiency in filtering redundant information while preserving task-relevant content.

MemAdapter achieves high memory utilization across all agent model scales, with utilization rates of 54.67\%, 56.68\%, and 61.88\% for the 1.5B, 3B, and 7B agent models, respectively. Notably, its advantage becomes more pronounced as model capacity increases, because memory utilization depends jointly on the quality of the memory system and the reasoning capability of the agent model. Most parametric and latent memory systems that require processing the full context exhibit relatively low utilization rates, reflecting the difficulty of extracting relevant information from long and redundant memory. MemoryLLM achieves relatively high utilization at smaller model scales, but its performance degrades as the agent model becomes more powerful. Overall, these results indicate that storing larger amounts of information does not necessarily lead to more reliable or effective memory usage by the agent model.

\begin{figure}[ht]
    \centering
    \includegraphics[width=\columnwidth]{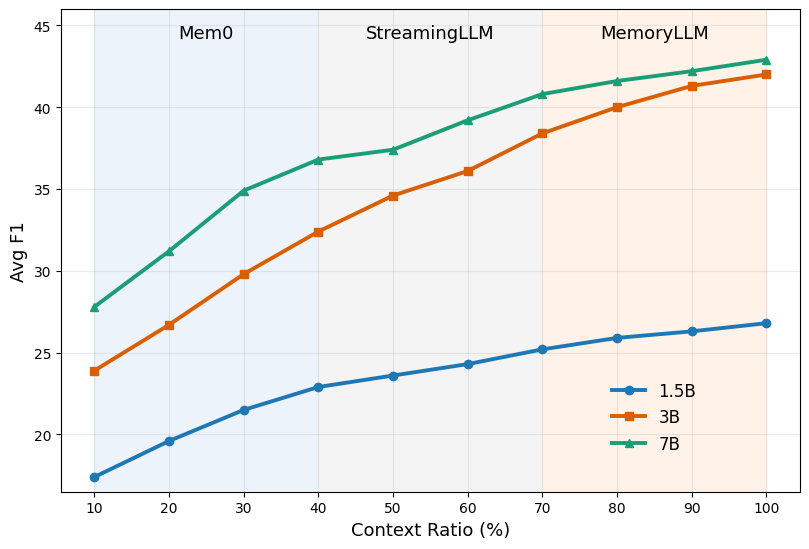}
    \caption{Increasing context information with heterogeneous memory encoding.
    Context segments at different positions are fused by Eqn.~\ref{eq:downsample}.}
    \label{fig:comlement_mem}
\end{figure}

\begin{figure*}[t!]
    \centering
    \includegraphics[width=0.95\textwidth]{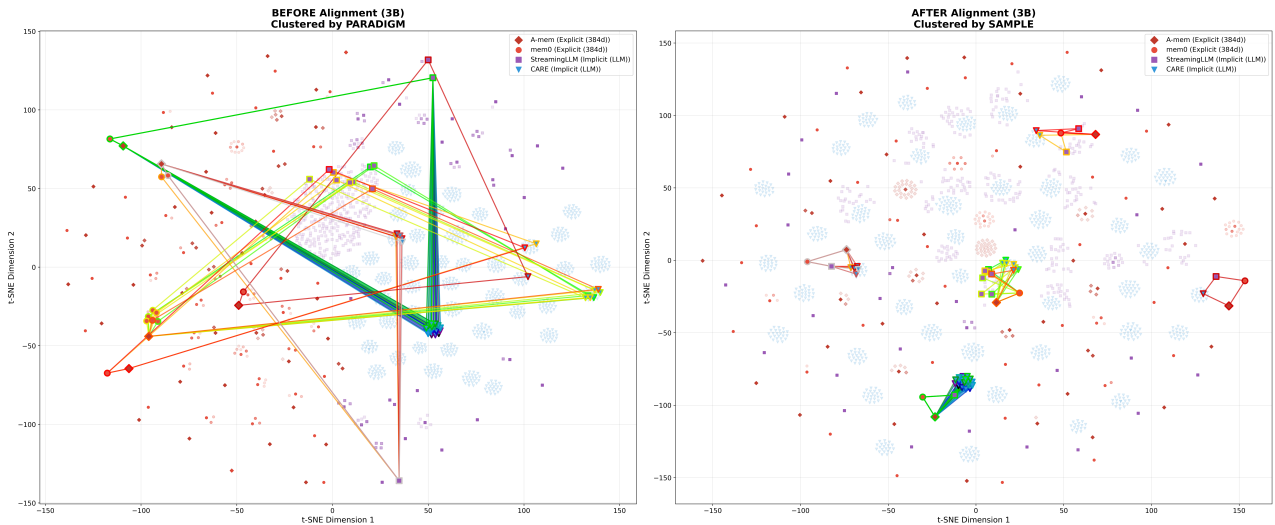}
    \caption{
    A t-SNE visualization of heterogeneous memory representations from the unified memory space, before and after cross-paradigm alignment. Memory states for parametric memory are obtained via the Qwen2.5-3B-Instruct agent model.
    }
    \label{fig:instance_alignment}
\end{figure*}

\subsection{Complemental Memory Fusion}

We further investigate the behavior of MemAdapter as the available context length increases under a heterogeneous memory setting.
The experiment is conducted on 1,000 randomly sampled queries. 
For each query, we reorganize the memory by encoding context segments at different positions using heterogeneous paradigms, including Mem0, StreamingLLM, and MemoryLLM. 
We then progressively increase the utilized context length from 10\% to 100\% of the original context and analyze the resulting changes in F1 performance, as illustrated in Fig.~\ref{fig:comlement_mem}.

The results show that the average F1 performance improves smoothly and monotonically as the memory ratio increases for all three agent models. This trend indicates that the proposed system can effectively integrate heterogeneous information from additional memory without introducing performance instability. Notably, the gains are distributed across the full range of memory ratios, rather than being concentrated in a specific interval. MemAdapter consistently translates incremental memory into performance improvements, suggesting that the aligned unified memory space enables reliable composition of heterogeneous evidence. This effect becomes more pronounced at larger model scales. While the 1.5B agent models exhibit steady but moderate gains, the 3B and 7B agent model benefits substantially from increased memory, achieving over 20 points improvement when the memory ratio increases from 10\% to 100\%. These results suggest that larger agent models can better exploit the high-quality, structured memory produced by MemAdapter, probably due to their more advanced in-context learning capabilities.

\subsection{Visualization and Analysis}
\subsubsection{Visualization of Unified Memory Space}
We visualize the effect of cross-paradigm alignment using t-SNE~\cite{maaten2008visualizing} to project heterogeneous memory representations from the unified memory space into the 2D space. 
The visualization is based on 1000 memory samples, encoded using multiple memory paradigms. For 35 samples, we connect their cross-paradigm representations with line segments to illustrate alignment behaviors, as shown in Fig.~\ref{fig:instance_alignment}.
Before cross-paradigm alignment (left), representations of the same sample are widely dispersed and form paradigm-specific clusters. In particular, explicit and implicit memory representations occupy distinct regions of the embedding space, leading to large cross-paradigm distances even for identical samples.
After alignment (right), representations of the same sample are consistently drawn closer, forming compact instance-level clusters despite originating from different paradigms. 
At the same time, representations of different samples remain distinguishable in the embedding space, indicating that the alignment preserves useful memory information rather than collapsing the representation space.
Overall, these results demonstrate that the proposed alignment method effectively reduces paradigm-induced discrepancies while maintaining instance-level discriminability in the unified memory space.

\begin{figure}[t]
    \centering
    \includegraphics[width=\columnwidth]{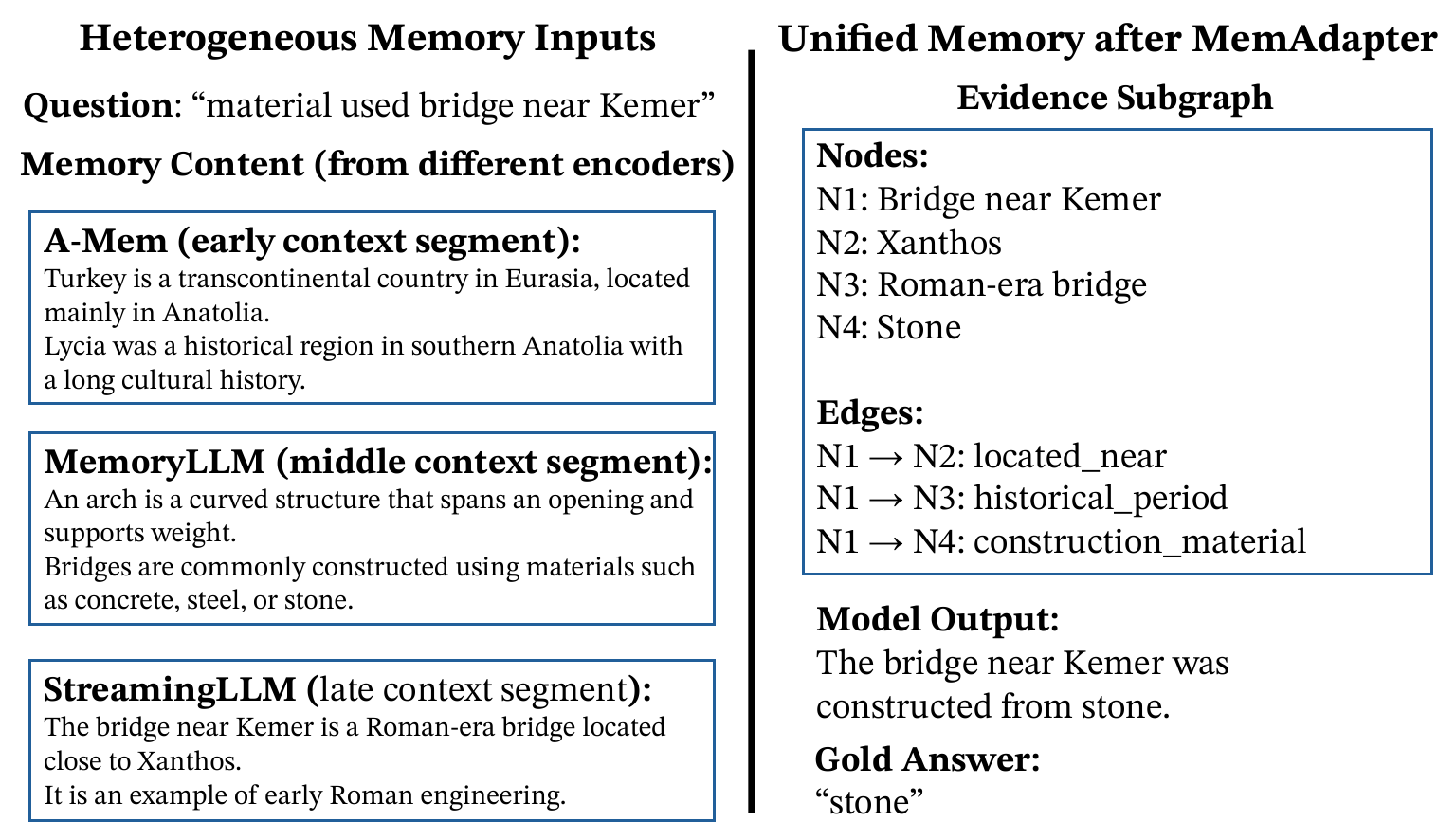}
    \caption{
    A case study of heterogeneous memory fusion.
    }
    \label{fig:memadapter_case_study}
\end{figure}

\subsubsection{Case Study}
Fig.~\ref{fig:memadapter_case_study} illustrates an example of how MemAdapter integrates heterogeneous memory into a unified evidence subgraph.
The input memories originate from non-overlapping context segments and are produced by heterogeneous memory systems.
In this example, none of the individual memory segments explicitly encodes the answer, and several segments contain background information that is only indirectly related to the question. As shown,
MemAdapter resolves the fragmentation by constructing an evidence subgraph that retains only entities and relations necessary for answering the question.
By aligning heterogeneous memory representations at the structural level, the model isolates the construction material relation of the target bridge while discarding irrelevant contextual content.
This case study demonstrates that effective memory utilization requires structured selection rather than surface-level aggregation, and highlights the role of heterogeneous memory fusion in enabling the agent model to perform accurate task execution. More cases are provided in Appendix \ref{appn:case_study}.

%% file: acm/Sections/experimental_setting.tex
\subsection{Training Details}
\label{sec:training_details}
\subsubsection{Baseline Methods}
We compare MemAdapter with five state-of-the-art agent memory systems that span explicit, parametric, and latent memory paradigms.
\textbf{Explicit memory}:
We include A-Mem~\cite{amem} and Mem0~\cite{mem0}, which store and retrieve explicit memory units (e.g., notes or memory entries) via embedding-based or graph-structured retrieval mechanisms.
\textbf{Parametric memory}:
We select StreamingLLM~\cite{streamingllm}, which enables efficient long-context inference through attention sink mechanisms without additional fine-tuning.
\textbf{Latent memory}:
We select MemoryLLM~\cite{memoryllm}, which maintains a trainable latent memory pool embedded within the Transformer layers, and CARE~\cite{care}, which encodes retrieved context into compact latent memory tokens via conflict-aware soft prompting.

\subsubsection{Models and Training Data}
During the model distillation process, the generative subgraph retriever is first trained using 40,000 instances sampled from HotpotQA~\cite{hotpotqa} training set (more details in Appendix~\ref{app:sampling}). The retriever is based on the Qwen2.5-1.5B~\cite{yang2025qwen3} model with LoRA-based~\cite{hu2022lora} fine-tuning, and the unified memory space dimension $\mathcal{D}_s=1536$, the same as the model's hidden state to facilitate implementation. The teacher model is instantiated using the Qwen2.5-32B-Instruct model. During the fast alignment stage, we use a fast alignment dataset consisting of merely \textbf{2,500} instances, including 1,200 samples from WikiMultiHopQA~\cite{wikimultihop} and 1,300 samples from NarrativeQA~\cite{narrativeqa} training set.
All performance on benchmarks is obtained by applying different memory systems on the Qwen2.5-instruct agent model series at three scales: 1.5B, 3B, and 7B, to study the performance of MemAdapter with increased in-context learning capacities of agent models.

\subsubsection{Evaluation Benchmarks}
We evaluate all memory systems on three publicly available multi-hop and memory-intensive question answering benchmarks: WikiMultiHopQA~\cite{wikimultihop}, NarrativeQA \cite{narrativeqa}, and MuSiQue~\cite{musique}, yielding 10,000, 10,557, and 2,417 questions respectively. Following prior work, we report standard QA metrics: accuracy by Exact Matching, F1, and ROUGE-1 measures to ensure consistent evaluation protocols across all baselines and model scales.

\subsubsection{Training Settings}
All experiments are conducted on NVIDIA H200 GPUs.
The generative subgraph retriever is trained on 6 GPUs and completes in approximately 1 hour.
The fast alignment process is conducted on a single GPU. Each new paradigm requires a memory state extraction process for about 10 minutes and 2.5 minutes for alignment module training, with a total of approximately 12-13 minutes.
We use AdamW optimization with learning rate $1\times10^{-5}$ for model distillation and $1\times10^{-4}$ for fast alignment. Alignment modules are implemented as light-weight feed-forward neural networks.
More details are provided in Appendix~\ref{app:hyperparams}.

%% file: acm/Tables/add_memadapter.tex
\definecolor{memgray}{RGB}{245,245,245}

\begin{table}[htbp]
\centering
\small
\setlength{\tabcolsep}{6pt}
\renewcommand{\arraystretch}{1.15}

\caption{
Fast alignment performance of MemAdapter.
$\uparrow$ and $\downarrow$ indicate comparisons to the original memory retrievers.
}
\label{tab:memadapter_effect}

\resizebox{\columnwidth}{!}{
\begin{tabular}{llcccc}
\toprule
\multirow{2}{*}{\textbf{Agent Model}} &
\multirow{2}{*}{\textbf{Method}} &
\multicolumn{2}{c}{\textbf{Original Retriever}} &
\multicolumn{2}{c}{\cellcolor{memgray}\textbf{MemAdapter}} \\
\cmidrule(lr){3-4}\cmidrule(lr){5-6}
& &
\textbf{F1} & \textbf{R-1} &
\cellcolor{memgray}\textbf{F1} & \cellcolor{memgray}\textbf{R-1} \\
\midrule

\multirow{3}{*}{Qwen2.5-1.5B}
& A-Mem       & 16.27 & 18.03 &
\cellcolor{memgray}19.08$\uparrow$ & \cellcolor{memgray}22.67$\uparrow$ \\
& MemoryLLM   & 13.88 & 15.02 &
\cellcolor{memgray}20.41$\uparrow$ & \cellcolor{memgray}23.96$\uparrow$ \\
& StreamingLLM & 18.92 & 21.36 &
\cellcolor{memgray}23.71$\uparrow$ & \cellcolor{memgray}26.84$\uparrow$ \\

\midrule

\multirow{3}{*}{Qwen2.5-3B}
& A-Mem       & 26.50 & 27.49 &
\cellcolor{memgray}27.91$\uparrow$ & \cellcolor{memgray}30.14$\uparrow$ \\
& MemoryLLM   & 22.72 & 23.78 &
\cellcolor{memgray}29.03$\uparrow$ & \cellcolor{memgray}31.88$\uparrow$ \\
& StreamingLLM & 28.25 & 29.37 &
\cellcolor{memgray}31.06$\uparrow$ & \cellcolor{memgray}33.92$\uparrow$ \\

\midrule

\multirow{3}{*}{Qwen2.5-7B}
& A-Mem       & 29.35 & 30.34 &
\cellcolor{memgray}33.18$\uparrow$ & \cellcolor{memgray}36.71$\uparrow$ \\
& MemoryLLM   & 28.05 & 29.17 &
\cellcolor{memgray}31.44$\uparrow$ & \cellcolor{memgray}35.02$\uparrow$ \\
& StreamingLLM & 36.28 & 37.35 &
\cellcolor{memgray}35.62$\downarrow$ & \cellcolor{memgray}41.04$\uparrow$ \\

\bottomrule
\end{tabular}
}

\end{table}

%% file: acm/Tables/memory_quality.tex



\begin{table}[ht]
\centering
\small
\setlength{\tabcolsep}{6pt}
\renewcommand{\arraystretch}{1.15}

\caption{
Comparisons on memory length, unique ratio of memory , and memory utilization across different systems.
}
\label{tab:memory_quality}

\resizebox{\columnwidth}{!}{
\begin{tabular}{l | cc | ccc}
\toprule
\multirow{2}{*}{\textbf{Method}} 
& \multicolumn{2}{c|}{\textbf{Memory Efficiency}} 
& \multicolumn{3}{c}{\textbf{Memory Utilization (\%) $\uparrow$}} \\
\cmidrule(lr){2-3} \cmidrule(lr){4-6}
& \textbf{Mem Length $\downarrow$} 
& \textbf{Unique Ratio (\%) $\uparrow$} 
& \textbf{1.5B} 
& \textbf{3B} 
& \textbf{7B} \\
\midrule
A-Mem 
& 4398.40 & 45.82 
& \textbf{56.50} & 51.24 & 50.45 \\
mem0 
& 4804.70 & 45.31 
& 51.73 & 55.93  & 53.68 \\
MemoryLLM 
& 5884.60 & -- 
& 48.94 & 54.16 & 59.29 \\
StreamingLLM 
& 5884.60 & --
& 44.68 & 48.20 & 49.83 \\
CARE 
& 5884.60 & --
& 40.39 & 35.70 & 42.03 \\
\midrule
\textbf{MemAdapter (Ours)} 
& \textbf{2241.80} & \textbf{46.92} 
& 54.67 & \textbf{56.68} & \textbf{61.88} \\
\bottomrule
\end{tabular}}
\end{table}

%% file: acm/Sections/related_work.tex
\section{Related Work}
Memory mechanisms in LLM-based agents can be broadly categorized into explicit, parametric, and latent paradigms, each with distinct representations and access mechanisms. Explicit memory systems store discrete records or structured graphs for retrieval, such as Reflexion, A-Mem, Mem0~\cite{NEURIPS2023_1b44b878,amem,mem0}. Parametric memory embeds information into model parameters or trainable modules, such as LMLM and MemoryLLM~\cite{lmlm,memoryllm}. Latent memory maintains evolving internal states, including KV caches or compressed prompts~\cite{streamingllm,zhang2023h2o}. Research in agent memory retrieval spans retrieval-augmented generation~\cite{DBLP:journals/corr/abs-2502-06872,DBLP:journals/corr/abs-2401-15391,edge2024local}, generative retrieval~\cite{DBLP:conf/nips/Tay00NBM000GSCM22,DBLP:conf/www/Zeng0JSWZ24,DBLP:journals/tois/LiJZZZZD25}, and subgraph retrieval~\cite{DBLP:conf/acl/ZhangZY000C22,DBLP:conf/acl/MavromatisK25,DBLP:journals/corr/abs-2506-00708}. More details in Appendix~\ref{appn:related_work}.

%% file: acm/Sections/conclusion.tex
\section{Conclusion}
In this work, we presented MemAdapter, a unified memory retrieval framework that enables fast alignment across heterogeneous agent memory paradigms. By combining a generative subgraph retriever with lightweight alignment modules, MemAdapter achieves cross-paradigm alignment within 13 minutes on a single GPU. Extensive experiments show that the generative subgraph retriever consistently outperforms strong memory paradigms, and the fast alignment module achieves performance superior to the original memory retrievers with less than 5\% of training
compute. MemAdapter also enables effective zero-shot fusion across memory paradigms, demonstrating its practicality as a plug-and-play solution for memory-augmented agent models.


%% file: acm/Sections/appn.tex
\section{Related Work}\label{appn:related_work}
\subsection{Agent Memory Paradigms}
Memory is a core design component in LLM-based agent systems, enabling stateful interaction and knowledge accumulation beyond a fixed context window~\cite{zhang2025survey,hu2025memory,jiaai,DBLP:journals/corr/abs-2504-15965}. Existing approaches can be grouped by how memories are represented and accessed.

\subsubsection{Explicit Memory}
Explicit memory stores past information as discrete records in external modules and retrieves them via similarity search or structured queries.
Reflexion~\cite{NEURIPS2023_1b44b878} maintains a buffer of self-reflections that are appended and reused to shape subsequent decisions.
A-Mem~\cite{amem} and Mem0~\cite{mem0} support long-term storage with embedding-based retrieval, optionally organizing records with explicit links for scalable access.
Graph-structured variants further model relations in memory, SGMem~\cite{wu2025sgmemsentencegraphmemory} encodes dialogue history as sentence-level graphs, and Zep~\cite{rasmussen2025zep} introduces a temporal knowledge-graph design with time-aware indexing.
AriGraph~\cite{DBLP:conf/ijcai/AnokhinSSEK0B25} combines episodic traces with an incrementally updated graph world model during environment interaction.

\subsubsection{Parametric Memory}
Parametric memory encodes information into model parameters or trainable internal modules, and is accessed implicitly through inference.
Limited Memory Language Models (LmLm)~\cite{lmlm} externalize entity-level facts to an external database during pre-training by interleaving lookup calls and masking retrieved values from the training loss, encouraging targeted retrieval rather than memorization.
MemoryLLM~\cite{memoryllm} introduces a fixed-size trainable memory pool inside transformer layers to support repeated memory writes.
Retroformer~\cite{yao2023retroformer} trains a retrospection component to generate feedback that can improve an actor model's behavior.
More generally, parameter-efficient adaptation methods (e.g., LoRA~\cite{hu2022lora}) and continual learning variants (e.g., sparse memory fine-tuning~\cite{lin2025continual}) provide mechanisms to incorporate new information while mitigating catastrophic forgetting~\cite{ramasesh2021effect}.

\subsubsection{Latent Memory}
Latent memory maintains information in model states that evolve during execution, such as KV caches or learned soft prompts.
StreamingLLM~\cite{streamingllm} enables efficient long-context decoding by combining attention sinks with a sliding-window KV cache.
H2O~\cite{zhang2023h2o} retains a subset of high-attention ``heavy-hitter'' tokens to compress the KV cache.
CARE~\cite{care} compresses retrieved context into soft-prompt tokens with conflict-aware training to improve robustness to noisy or contradictory evidence.
MemGen~\cite{zhang2025memgen} proposes generative latent memory for self-evolving agents, updating latent representations as new information arrives.

These lines of work typically assume a memory representation and access mechanism tied to a specific paradigm, which makes combining heterogeneous memories nontrivial.

\subsection{Agent Memory Retrieval}
\subsubsection{Retrieval-augmented Generation}
Retrieval-augmented generation (RAG) grounds LLM outputs in external evidence~\cite{DBLP:journals/corr/abs-2502-06872}.
Standard RAG retrieves independent passages via embedding similarity, which can be insufficient for queries requiring multi-hop evidence aggregation~\cite{DBLP:journals/corr/abs-2401-15391}.
Graph-based RAG methods leverage relational structure to retrieve and organize connected evidence~\cite{DBLP:journals/corr/abs-2408-08921}.
GraphRAG~\cite{edge2024local}, for example, summarizes entity graphs via community structure to support global queries, while RAPTOR~\cite{DBLP:conf/iclr/SarthiATKGM24} builds a hierarchical tree of summaries to enable multi-resolution retrieval.
However, these methods primarily operate over explicit text stores and do not directly address heterogeneous agent memory paradigms.

\subsubsection{Generative Retrieval}
Generative retrieval formulates retrieval as sequence generation rather than ranking.
DSI~\cite{DBLP:conf/nips/Tay00NBM000GSCM22} autoregressively generates document identifiers from a query, establishing a foundational approach to corpus indexing via model parameters.
GENRE~\cite{DBLP:conf/iclr/CaoI0P21} retrieves entities by generating names with constrained decoding.
Subsequent work improves scalability through identifier design and training strategies~\cite{DBLP:conf/www/Zeng0JSWZ24,DBLP:journals/tois/LiJZZZZD25}.

\subsubsection{Subgraph Retrieval}
Subgraph retrieval seeks to extract task-relevant substructures from knowledge graphs to support downstream reasoning.
SR~\cite{DBLP:conf/acl/ZhangZY000C22} introduces a trainable subgraph retriever that is decoupled from the reasoning module, yielding a plug-and-play framework applicable to any subgraph-oriented QA model.
GNN-RAG~\cite{DBLP:conf/acl/MavromatisK25} employs GNNs as dense subgraph reasoners to extract candidate answer paths, which are then verbalized for LLM reasoning. Notably, GNN-RAG achieves performance competitive with GPT-4 on multi-hop KGQA benchmarks using only a 7B model.
GRAG~\cite{DBLP:journals/corr/abs-2408-08921} proposes a divide-and-conquer strategy for efficient textual subgraph retrieval and integrates the retrieved structure through dual text-graph views.
DrKGC~\cite{DBLP:journals/corr/abs-2506-00708} combines rule-guided bottom-up subgraph retrieval with a GCN adapter for knowledge graph completion.

\subsubsection{Contrastive Learning}
Contrastive learning has proven effective for aligning heterogeneous representation spaces.
CLIP~\cite{DBLP:conf/icml/RadfordKHRGASAM21} shows that contrastive pre-training on image-text pairs produces a shared embedding space that supports zero-shot cross-modal transfer, with its InfoNCE-style objective becoming a widely adopted alignment technique.
SimCLR~\cite{chen2020simple} and SimCSE~\cite{gao2021simcse} establish contrastive learning frameworks for visual and textual representations, respectively.
In the retrieval domain, progressive distillation~\cite{DBLP:conf/www/LinGLZLD0LJMD23} addresses the capacity gap between teacher and student retrievers through staged knowledge transfer with increasingly capable teachers.

\section{Additional Implementation Details}
\label{sec:appendix_impl}

\subsection{Two Step Teacher Generation}
\label{sec:appendix_teacher}

Stage~I supervision is generated by a two step teacher procedure. In the first step, the teacher constructs a query independent full memory graph $\mathcal{G}^{\star}$ from a document collection. The objective of this step is coverage rather than minimality. Redundant nodes and edges are allowed as long as they reflect information present in the documents. In the second step, given a query and the fixed full graph, the teacher extracts an evidence subgraph $r^{\star}$ that is constrained to be a subgraph of $\mathcal{G}^{\star}$. This subset constraint enforces structural consistency between the anchored memory state and the evidence supervision and enables deterministic verification.

\subsection{Prompt Templates for Teacher Generation}
\label{sec:appendix_prompt}

We use strict prompting templates to enforce a fixed schema for both full graph construction and evidence subgraph extraction. The templates are shown below.

\begin{figure}[t]
\centering
\setlength{\fboxsep}{6pt}
\fcolorbox{blue!50!black}{blue!3}{
\begin{minipage}{0.92\linewidth}
\small
\textbf{Step 1. Full Graph Construction Prompt (Strict).}

\vspace{0.3em}
\textbf{TASK:} Create a FULL memory graph covering all entities and relations in the documents.

\vspace{0.3em}
\textbf{RULES:}
(1) Include all important entities and do not filter for any specific question.
(2) Use node IDs N1, N2, N3, and so on.
(3) List all nodes before any edges.
(4) Follow the required output format.

\vspace{0.3em}
\textbf{OUTPUT FORMAT (REQUIRED):}
\texttt{[FULL\_GRAPH]} \\
\texttt{<NODES>} \\
\texttt{N1: description} \\
\texttt{...} \\
\texttt{<EDGES>} \\
\texttt{N1 -> N2: relation} \\
\texttt{...}
\end{minipage}
}
\caption{Strict prompt for query independent full graph construction.}
\label{fig:prompt_step1}
\end{figure}

\begin{figure}[t]
\centering
\setlength{\fboxsep}{6pt}
\fcolorbox{orange!70!black}{orange!3}{
\begin{minipage}{0.92\linewidth}
\small
\textbf{Step 2. Evidence Subgraph Extraction Prompt (Strict).}

\vspace{0.3em}
\textbf{TASK:} From the full graph, extract nodes and edges that help answer the question.

\vspace{0.3em}
\textbf{CRITICAL RULES:}
(1) All nodes must exist in the full graph with the same ID and description.
(2) All edges must exist in the full graph with the same source, target, and relation.
(3) Include complete node information and complete edge information.
(4) Follow the required output format.

\vspace{0.3em}
\textbf{OUTPUT FORMAT (REQUIRED):}
\texttt{[EVIDENCE\_SUBGRAPH]} \\
\texttt{<NODES>} \\
\texttt{N1: description} \\
\texttt{...} \\
\texttt{<EDGES>} \\
\texttt{N1 -> N2: relation} \\
\texttt{...} \\
\texttt{[CONFIDENCE]} \\
\texttt{0.85}

\vspace{0.3em}
\textbf{VERIFICATION CHECKLIST:} The prompt includes a checklist to verify format compliance and the subset constraint.

\vspace{0.3em}
\textbf{WARNING:} If any node or edge does not exist in the full graph, the output is rejected.
\end{minipage}
}
\caption{Strict prompt for evidence subgraph extraction with subset verification and confidence reporting.}
\label{fig:prompt_step2}
\end{figure}

\subsection{Stage-1 Data Sampling and Training Details}
\label{app:sampling}
To train the anchor memory representation in Stage-1, we construct a teacher-supervised training set from HotpotQA~\cite{hotpotqa} with graph-structured evidence annotations.
Rather than using the full training split, we select a subset of \textbf{40,000} instances to balance training efficiency and representation quality.

The subset is constructed using a controlled sampling strategy that considers multiple factors jointly.
Specifically, each candidate instance is assigned a composite sampling score based on:
(i) input length,
(ii) reasoning complexity measured by the number of supporting facts,
(iii) structural diversity of the associated evidence graph, and
(iv) a randomized component to preserve coverage of the original data distribution.
Sampling is performed with a fixed random seed to ensure reproducibility.

This strategy avoids over-representing extremely long or trivial instances while maintaining sufficient diversity for learning a stable anchor representation space.
All sampled instances retain their original teacher-provided graph annotations, which are used as supervision signals during training.
Stage-1 training is conducted on the Qwen2.5-1.5B base model using LoRA-based fine-tuning.
Only the parameters of MemAdapter and the anchor-side alignment module are updated, while the agent model remains frozen.



\subsection{Full Hyperparameter Settings}
\label{app:hyperparams}

Table~\ref{tab:stage1_hyperparams} and Table~\ref{tab:stage2_hyperparams} list the hyperparameters used for Stage~1 (model distillation) and Stage~2 (fast alignment). Inference settings for evaluation across agent model scales (1.5B, 3B, 7B) are summarized in Table~\ref{tab:inference_hyperparams}.

\begin{table*}[]
\centering
\caption{Inference (evaluation) settings. All memory systems are evaluated with the same reasoner model series (Qwen2.5-instruct 1.5B, 3B, 7B) on the merged test set (22,974 questions).}
\label{tab:inference_hyperparams}
\begin{tabular}{ll}
\toprule
\textbf{Setting} & \textbf{Value} \\
\midrule
Reasoner models & Qwen2.5-1.5B-Instruct, 3B-Instruct, 7B-Instruct \\
Inference batch size & 8 (or 16--32 where memory allows) \\
Max new tokens (answer) & 128--256 \\
Decoding & Greedy or sampling (consistent across baselines) \\
Test set & 22,974 (WikiMultiHopQA 10k + NarrativeQA 10,557 + MuSiQue 2,417) \\
\bottomrule
\end{tabular}
\end{table*}

All experiments use a fixed random seed (42) for reproducibility. Stage~1 and Stage~2 implementation details and data preparation scripts are available in the supplementary code.

\begin{figure*}[htbp]
    \centering
    \includegraphics[width=0.49\textwidth]{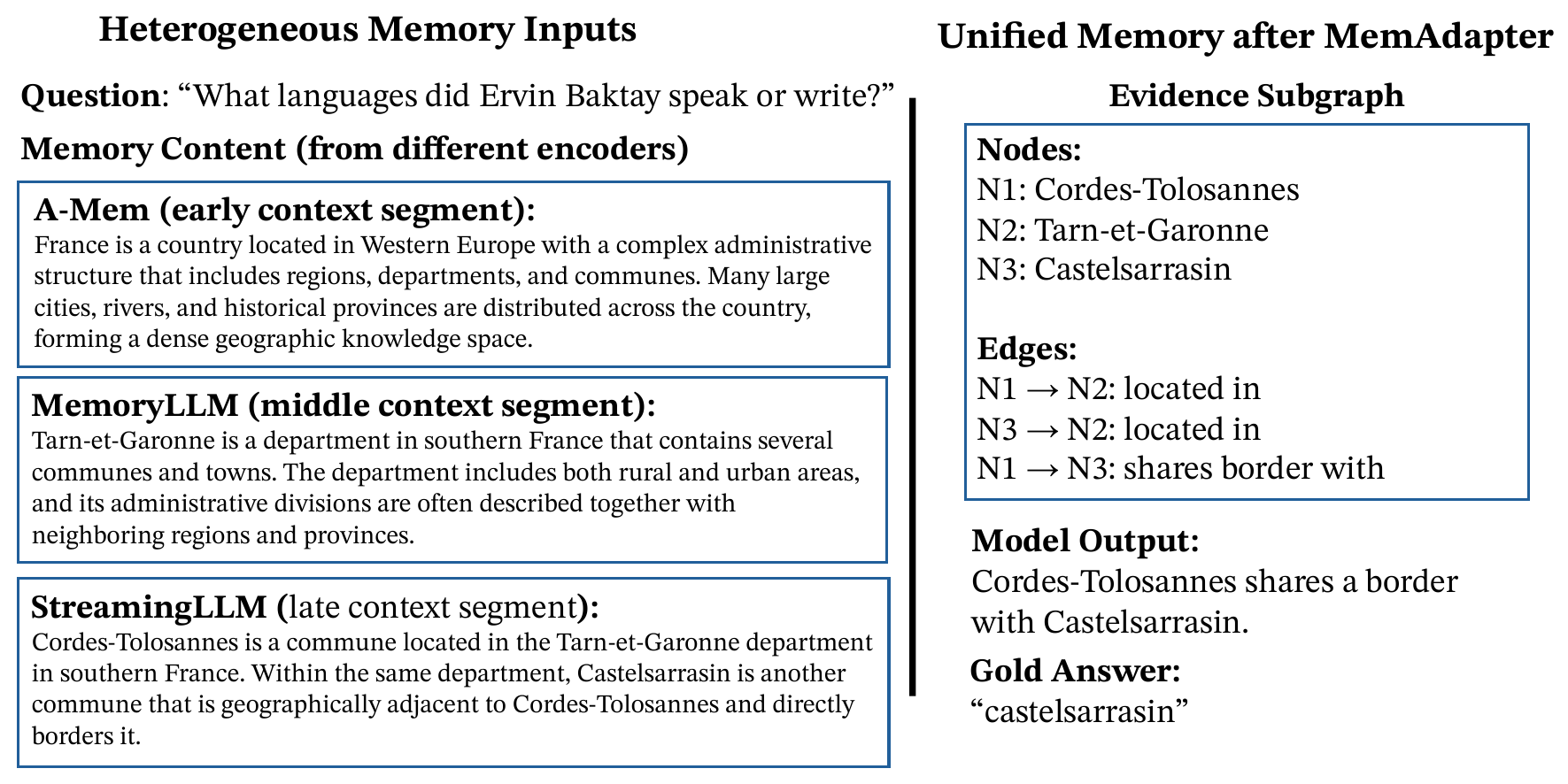}
    \hfill
    \includegraphics[width=0.49\textwidth]{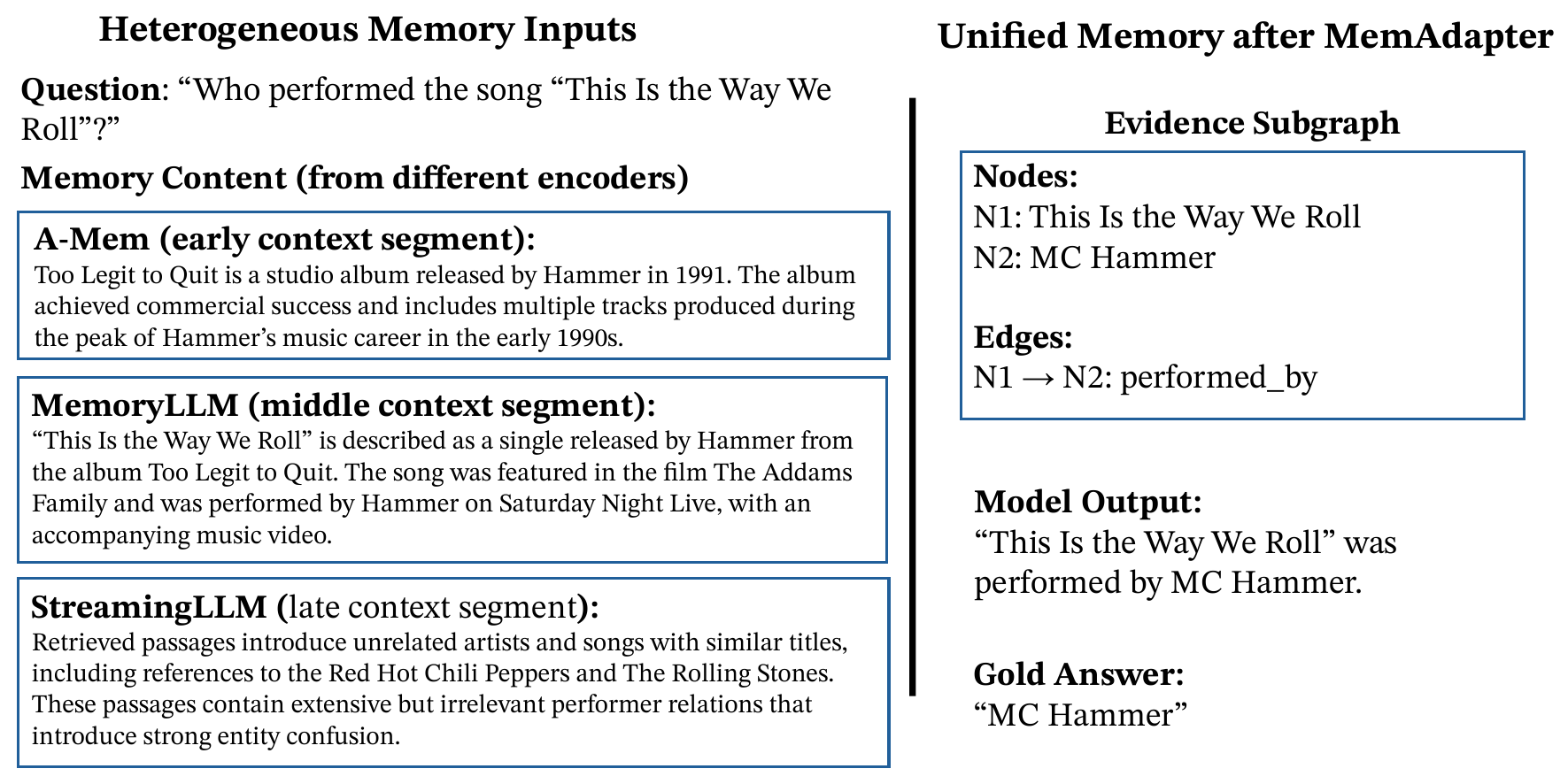}
    \caption{
    Additional case studies illustrating memory unification via evidence subgraph construction.
    Each example presents heterogeneous memory inputs produced by different encoders from disjoint context segments, followed by the unified evidence subgraph selected by MemAdapter and the resulting model output.
    }
    \label{fig:appendix_case_studies}
\end{figure*}

\begin{table*}[htbp]
\centering
\caption{Stage~1 (model distillation) hyperparameters. The student model is Qwen2.5-1.5B with LoRA; training uses KL distillation.}
\label{tab:stage1_hyperparams}
\begin{tabular}{ll}
\toprule
\textbf{Setting} & \textbf{Value} \\
\midrule
Training data & 40,000 (HotpotQA, see Appendix~\ref{app:sampling}) \\
Train/val split & 90\% / 10\% (seed 42) \\
Per-device batch size & 4 \\
Gradient accumulation steps & 4 \\
Effective batch size & $4 \times N_{\text{GPUs}} \times 4$ (e.g., 128 for 8 GPUs) \\
Epochs & 3 \\
Learning rate & $1\times10^{-5}$ \\
Warmup & (as in optimizer; typically 5\% of steps) \\
Optimizer & AdamW \\
Weight decay & 0.01 \\
Max input length & 4096 tokens \\
Max output length & 512 tokens \\
Precision & BF16 \\
\midrule
\multicolumn{2}{l}{\textit{LoRA}} \\
\quad Rank $r$ & 16 \\
\quad Alpha & 32 \\
\quad Dropout & 0.1 \\
\quad Target modules & \texttt{q\_proj}, \texttt{v\_proj} \\
\midrule
\multicolumn{2}{l}{\textit{Distillation (KL + CE)}} \\
\quad KL weight & 0.5 \\
\quad KL temperature & 2.0 \\
\quad CE weight & 1.0 \\
\bottomrule
\end{tabular}
\end{table*}

\begin{table*}[]
\centering
\caption{Stage~2 (fast alignment) hyperparameters. Alignment is trained per paradigm (A-Mem, StreamingLLM, MemoryLLM) and per agent model scale (1.5B, 3B, 7B); anchor dimension is 1536.}
\label{tab:stage2_hyperparams}
\begin{tabular}{ll}
\toprule
\textbf{Setting} & \textbf{Value} \\
\midrule
Alignment data & 2,500 (1,200 WikiMultiHopQA + 1,300 NarrativeQA) \\
Batch size & 32 \\
Epochs & 10--20 (we use 20 in main runs) \\
Learning rate & $1\times10^{-4}$ \\
Weight decay & 0.01 \\
Warmup ratio & 0.1 \\
Optimizer & AdamW \\
Scheduler & Cosine annealing \\
\midrule
\multicolumn{2}{l}{\textit{Loss}} \\
\quad Contrastive (InfoNCE) temperature & 0.07 \\
\quad MSE loss (optional) & enabled, weight 0.1 \\
\quad Anchor module & frozen \\
\midrule
\multicolumn{2}{l}{\textit{Model scale $\rightarrow$ hidden dim}} \\
\quad 1.5B & 1536 \\
\quad 3B & 2048 \\
\quad 7B & 3584 \\
\bottomrule
\end{tabular}
\end{table*}

\section{Supplementary Case Study: Evidence Subgraph Construction}
\label{appn:case_study}
The two examples in Figure~\ref{fig:appendix_case_studies} highlight different sources of difficulty when reasoning over heterogeneous memory inputs.
In the first case, the challenge arises from severe entity and relation confusion, where multiple retrieved memories introduce competing performer relations for similar or identical song titles.
In the second case, the difficulty stems from heavy structural noise, where large amounts of geographic and administrative information obscure a simple local adjacency relation.

In both scenarios, MemAdapter constructs a compact evidence subgraph that preserves only the entities and relations required for answering the query, while discarding memory content that is either redundant or structurally irrelevant.
These examples illustrate that effective memory utilization is not achieved through aggregating more content, but through enforcing structural consistency and strict subset selection.
As a result, MemAdapter enables reliable reasoning even when memory inputs are heterogeneous, incomplete, or weakly aligned with the query.